# INTRUSION DETECTION SYSTEMS USING ADAPTIVE REGRESSION SPLINES

Srinivas Mukkamala, Andrew H. Sung
*Department of Computer Science, New Mexico Tech, Socorro, U.S.A.*
*srinivas@cs.nmt.edu, sung@cs.nmt.edu*

Ajith Abraham
*Natural Computation Lab. Dep. Computer Science, Oklahoma State University, Tulsa, U.S.A.*
*ajith.abraham@ieee.org*

Vitorino Ramos
*CVRM-IST, Instituto Superior Técnico,Technical University of Lisbon, Lisbon, Portugal*
*vitorino.ramos@alfa.ist.utl.pt*

Keywords: Network security, intrusion detection, adaptive regression splines, neural networks, support vector machines

Abstract: Past few years have witnessed a growing recognition of intelligent techniques for the construction of efficient and reliable intrusion detection systems. Due to increasing incidents of cyber attacks, building effective intrusion detection systems (IDS) are essential for protecting information systems security, and yet it remains an elusive goal and a great challenge. In this paper, we report a performance analysis between Multivariate Adaptive Regression Splines (MARS), neural networks and support vector machines. The MARS procedure builds flexible regression models by fitting separate splines to distinct intervals of the predictor variables. A brief comparison of different neural network learning algorithms is also given.

## 1 INTRODUCTION

Intrusion detection is a problem of great significance to protecting information systems security, especially in view of the worldwide increasing incidents of cyber attacks. Since the ability of an IDS to classify a large variety of intrusions in real time with accurate results is important, we will consider performance measures in three critical aspects: training and testing times; scalability; and classification accuracy.

Since most of the intrusions can be located by examining patterns of user activities and audit records (Denning, 1987), many IDSs have been built by utilizing the recognized attack and misuse patterns. IDSs are classified, based on their functionality, as misuse detectors and anomaly detectors. Misuse detection systems use well-known attack patterns as the basis for detection (Denning, 1987; Kumar, 1994). Anomaly detection systems use user profiles as the basis for detection; any deviation from the normal user behaviour is considered an intrusion (Denning, 1987; Kumar, 1994; Ghosh, 1999; Cannady, 1998).

One of the main problems with IDSs is the overhead, which can become unacceptably high. To analyse system logs, the operating system must keep information regarding all the actions performed, which invariably results in huge amounts of data, requiring disk space and CPU resource.

Next, the logs must be processed to convert into a manageable format and then compared with the set of recognized misuse and attack patterns to identify possible security violations. Further, the stored patterns need be continually updated, which would normally involve human expertise. An intelligent, adaptable and cost-effective tool that is capable of (mostly) real-time intrusion detection is the goal of the researchers in IDSs. Various artificial intelligence techniques have been utilized to automate the intrusion detection process to reduce human intervention; several such techniques include neural networks (Ghosh, 1999; Cannady, 1998; Ryan 1998; Debar, 1992a-b), and machine learning (Mukkamala, 2002a). Several data mining techniques have been introduced to identify key features or parameters that define intrusions (Luo, 2000; Cramer, 1995; Stolfo, 2000 ,Mukkamala, 2002b).

In this paper, we explore Multivariate Adaptive Regression Splines (MARS), Support Vector Machines (SVM) and Artificial Neural Networks (ANN), to perform intrusion detection based on recognized attack patterns. The data we used in our experiments originated from MIT's Lincoln Lab. It was developed for intrusion detection system evaluations by DARPA and is considered a benchmark for IDS evaluations (Lincoln Laboratory, 1998-2000).

We perform experiments to classify the network traffic patterns according to a 5-class taxonomy. The five classes of patterns in the DARPA data are (normal, probe, denial of service, user to super-user, and remote to local).

In the rest of the paper, a brief introduction to the data we use is given in section 2. Section 3 briefly introduces to MARS. In section 4 a brief introduction to the connectionist paradigms (ANNs and SVMs) is given. In section 5 the experimental results of using MARS, ANNs and SVMs are given. The summary and conclusions of our work are given in section 6.

## 2 INTRUSION DETECTION DATA

In the 1998 DARPA intrusion detection evaluation program, an environment was set up to acquire raw TCP/IP dump data for a network by simulating a typical U.S. Air Force LAN. The LAN was operated like a real environment, but being blasted with multiple attacks (Kris, 1998; Seth, 1998). For each TCP/IP connection, 41 various quantitative and qualitative features were extracted (Stolfo, 2000; University of California at Irvine, 1999). Of this database a subset of 494021 data were used, of which 20% represent normal patterns. Attack types fall into four main categories:

- Probing: surveillance and other probing
- DoS: denial of service
- U2Su: unauthorized access to local super user (root) privileges
- R2L: unauthorized access from a remote machine.

### 2.1 Probing

Probing is a class of attacks where an attacker scans a network to gather information or find known vulnerabilities. An attacker with a map of machines and services that are available on a network can use the information to look for exploits. There are different types of probes: some of them abuse the computer's legitimate features; some of them use social engineering techniques. This class of attacks is the most commonly heard and requires very little technical expertise.

### 2.2 Denial of Service Attacks

Denial of Service (DoS) is a class of attacks where an attacker makes some computing or memory resource too busy or too full to handle legitimate requests, thus denying legitimate users access to a machine. There are different ways to launch DoS attacks: by abusing the computers legitimate features; by targeting the implementations bugs; or by exploiting the system's misconfigurations. DoS attacks are classified based on the services that an attacker renders unavailable to legitimate users.

### 2.3 User to Root Attacks

User to root exploits are a class of attacks where an attacker starts out with access to a normal user account on the system and is able to exploit vulnerability to gain root access to the system. Most common exploits in this class of attacks are regular buffer overflows, which are caused by regular programming mistakes and environment assumptions.

### 2.4 Remote to User Attacks

A remote to user (R2L) attack is a class of attacks where an attacker sends packets to a machine over a network, then exploits machine's vulnerability to illegally gain local access as a user. There are different types of R2U attacks; the most common attack in this class is done using social engineering.

# 3 MULTIVARIATE ADAPTIVE REGRESSION SPLINES (MARS)

Splines can be considered as an innovative mathematical process for complicated curve drawings and function approximation. To develop a spline the X-axis is broken into a convenient number of regions. The boundary between regions is also known as a knot. With a sufficiently large number of knots virtually any shape can be well approximated. While it is easy to draw a spline in 2-dimensions by keying on knot locations (approximating using linear, quadratic or cubic polynomial etc.), manipulating the mathematics in higher dimensions is best accomplished using basis functions. The MARS model is a regression model using basis functions as predictors in place of the original data. The basis function transform makes it possible to selectively blank out certain regions of a variable by making them zero, and allows MARS to focus on specific sub-regions of the data. It excels at finding optimal variable transformations and interactions, and the complex data structure that often hides in high-dimensional data (Friedman, 1991).

Given the number of records in most data sets, it is infeasible to approximate the function $y=f(x)$ by summarizing $y$ in each distinct region of $x$. For some variables, two regions may not be enough to track the specifics of the function. If the relationship of $y$ to some $x's$ is different in 3 or 4 regions, for example, the number of regions requiring examination is even larger than 34 billion with only 35 variables. Given that the number of regions cannot be specified a priori, specifying too few regions in advance can have serious implications for the final model. A solution is needed that accomplishes the following two criteria:

- judicious selection of which regions to look at and their boundaries
- judicious determination of how many intervals are needed for each variable.

Given these two criteria, a successful method will essentially need to be adaptive to the characteristics of the data. Such a solution will probably ignore quite a few variables (affecting variable selection) and will take into account only a few variables at a time (also reducing the number of regions). Even if the method selects 30 variables for the model, it will not look at all 30 simultaneously. Such simplification is accomplished by a decision tree at a single node, only ancestor splits are being considered; thus, at a depth of six levels in the tree, only six variables are being used to define the node.

## 3.1 MARS Smoothing, Splines, Knots Selection and Basis Functions

To estimate the most common form, the cubic spline, a uniform grid is placed on the predictors and a reasonable number of knots are selected. A cubic regression is then fit within each region. This approach, popular with physicists and engineers who want continuous second derivatives, requires many coefficients (four per region), in order to be estimated. Normally, two constraints, which dramatically reduce the number of free parameters, can be placed on cubic splines: curve segments must join, and continuous first and second derivatives at knots (higher degree of smoothness).

Figure 1 shows typical attacks and their distribution while figure 2 (section 5) depicts a MARS spline with three knots (actual data on the rigth). A key concept underlying the spline is the knot. A knot marks the end of one region of data and the beginning of another. Thus, the knot is where the behavior of the function changes. Between knots, the model could be global (e.g., linear regression). In a classical spline, the knots are predetermined and evenly spaced, whereas in MARS, the knots are determined by a search procedure. Only as many knots as needed are included in a MARS model. If a straight line is a good fit, there will be no interior knots. In MARS, however, there is always at least one "pseudo" knot that corresponds to the smallest observed value of the predictor (Steinberg, 1999). Finding the one best knot in a simple regression is a straightforward search problem: simply examine a large number of potential knots and choose the one with the best $R^2$. However, finding the best pair of knots requires far more computation, and finding the best set of knots when the actual number needed is unknown is an even more challenging task. MARS finds the location and number of needed knots in a forward/backward stepwise fashion. A model which is clearly over fit with too many knots is generated first; then, those knots that contribute least to the overall fit are removed. Thus, the forward knot selection will include many incorrect knot locations, but these erroneous knots will eventually (although this is not guaranteed), be deleted from the model in the backwards pruning step (Abraham, 2001).

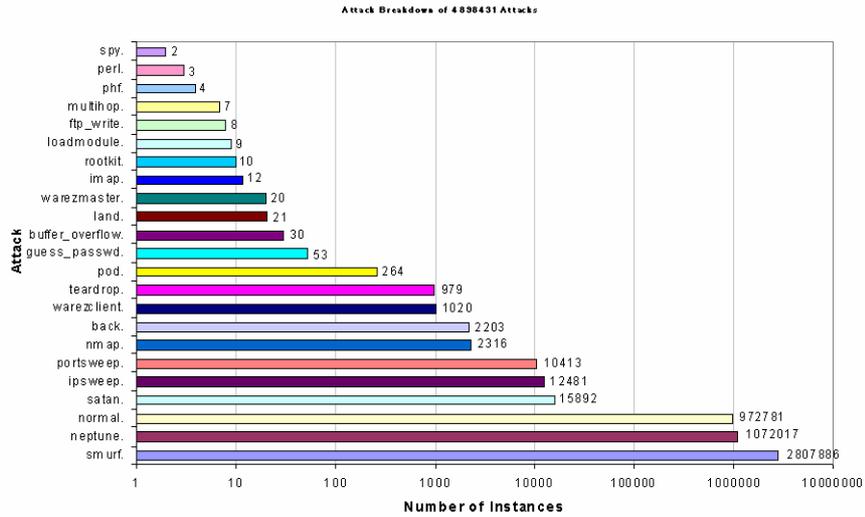

Figure1. Intrusion Detection Data Distribution

## 4 CONNECTIONIST PARADIGMS

The artificial neural network (ANN) methodology enables us to design useful nonlinear systems accepting large numbers of inputs, with the design based solely on instances of input-output relationships.

### 4.1 Resilient Back propagation (RP)

The purpose of the resilient back propagation training algorithm is to eliminate the harmful effects of the magnitudes of the partial derivatives. Only the sign of the derivative is used to determine the direction of the weight update; the magnitude of the derivative has no effect on the weight update. The size of the weight change is determined by a separate update value. The update value for each weight and bias is increased by a factor whenever the derivative of the performance function with respect to that weight has the same sign for two successive iterations. The update value is decreased by a factor whenever the derivative with respect that weight changes sign from the previous iteration. If the derivative is zero, then the update value remains the same. Whenever the weights are oscillating the weight change will be reduced. If the weight continues to change in the same direction for several iterations, then the magnitude of the weight change will be increased (Riedmiller, 1993).

### 4.2 Scaled Conjugate Gradient Algorithm (SCG)

The scaled conjugate gradient algorithm is an implementation of avoiding the complicated line search procedure of conventional conjugate gradient algorithm (CGA). According to the SCGA, the Hessian matrix is approximated by

$$E''(w_k)p_k = \frac{E'(w_k + s_k p_k) - E'(w_k)}{s_k} + \lambda_k p_k$$

where $E'$ and $E''$ are the first and second derivative information of global error function $E(w_k)$. The other terms $p_k$, $s_k$ and $\lambda_k$ represent the weights, search direction, parameter controlling the change in weight for second derivative approximation and parameter for regulating the indefiniteness of the Hessian. In order to get a good quadratic approximation of $E$, a mechanism to raise and lower $\lambda_k$ is needed when the Hessian is positive definite (Moller, 1993).

### 4.3 One-Step-Secant Algorithm (OSS)

Quasi-Newton method involves generating a sequence of matrices $G^{(k)}$ that represents increasingly accurate approximations to the inverse Hessian $(H^{-1})$. Using only the first derivative information of $E$ the updated expression is as follows:

$$G^{(k+1)} = G^{(k)} + \frac{pp^T}{p^T v} - \frac{(G^{(k)}v)v^T G^{(k)}}{v^T G^{(k)} v} + (v^T G^{(k)} v)uu^T$$

where
$$p = w^{(k+1)} - w^{(k)}, \quad v = g^{(k+1)} - g^{(k)},$$
$$u = \frac{p}{p^T v} - \frac{G^{(k)} v}{v^T G^{(k)} v}$$

and *T* represents transpose of a matrix. The problem with this approach is the requirement of computation and storage of the approximate Hessian matrix for every iteration. The One-Step-Secant (OSS) is an approach to bridge the gap between the conjugate gradient algorithm and the quasi-Newton (secant) approach. The OSS approach doesn't store the complete Hessian matrix; it assumes that at each iteration the previous Hessian was the identity matrix. This also has the advantage that the new search direction can be calculated without computing a matrix inverse (Bishop, 1995).

### 4.4 Support Vector Machines (SVMs)

The SVM approach transforms data into a feature space *F* that usually has a huge dimension. It is interesting to note that SVM generalization depends on the geometrical characteristics of the training data, not on the dimensions of the input space (Bishop, 1995; Joachims, 1998). Training a support vector machine (SVM) leads to a quadratic optimization problem with bound constraints and one linear equality constraint. Vapnik (Vladimir, 1995) shows how training a SVM for the pattern recognition problem leads to the following quadratic optimization problem (Joachims, 2000):

Minimize:
$$W(\mathbf{a}) = -\sum_{i=1}^{l} \mathbf{a}_i + \frac{1}{2} \sum_{i=1}^{l} \sum_{j=1}^{l} y_i y_j \mathbf{a}_i \mathbf{a}_j k(x_i, x_j) \quad (1)$$

Subject to
$$\sum_{i=1}^{l} y_i \mathbf{a}_i \quad (2)$$
$$\forall i: 0 \leq \mathbf{a}_i \leq C$$

where *l* is the number of training examples **a** is a vector of *l* variables and each component $\mathbf{a}_i$ corresponds to a training example $(x_i, y_i)$. The solution of (1) is the vector $\mathbf{a}^*$ for which (1) is minimized and (2) is fulfilled.

## 5 EXPERIMENTS

In our experiments, we perform 5-class classification. The (training and testing) data set contains 11982 randomly generated points from the data set representing the five classes, with the number of data from each class proportional to its size, except that the smallest classes are completely included. The normal data belongs to class1, probe belongs to class 2, denial of service belongs to class 3, user to super user belongs to class 4, remote to local belongs to class 5. A different randomly selected set of 6890 points of the total data set (11982) is used for testing MARS, SVMs and ANNs. Overall accuracy of the classifiers is given in Tables 1-4. Class specific classification of the classifiers is given in Table 5.

### 5.1 MARS Experiments

We used 5 basis functions and selected a setting of minimum observation between knots as 10. The MARS training mode is being set to the lowest level to gain higher accuracy rates. Five MARS models are employed to perform five class classifications (normal, probe, denial of service, user to root and remote to local). We partition the data into the two classes of "Normal" and "Rest" (Probe, DoS, U2Su, R2L) patterns, where the Rest is the collection of four classes of attack instances in the data set. The objective is to separate normal and attack patterns. We repeat this process for all classes. Table 1 summarizes the results of the experiments.

### 5.2 Neural Network Experiments

The same data set described in section 2 is being used for training and testing different neural network algorithms. The set of 5092 training data is divided in to five classes: normal, probe, denial of service attacks, user to super user and remote to local attacks. Where the attack is a collection of 22 different types of instances that belong to the four classes described in section 2, and the other is the normal data. In our study we used two hidden layers with 20 and 30 neurons each and the networks were trained using training functions described in Table 6. The network was set to train until the desired mean square error of 0.001 was met. As multi-layer feed forward networks are capable of multi-class classifications, we partition the data into 5 classes (Normal, Probe, Denial of Service, and User to Root and Remote to Local).

Table 1. MARS Test Performance

| Class | Accuracy |
|---|---|
| Normal | 96.08 % |
| Probe | 92.32 % |
| DOS | 94.73 % |
| U2Su | 99.71 % |
| R2L | 99.48 % |

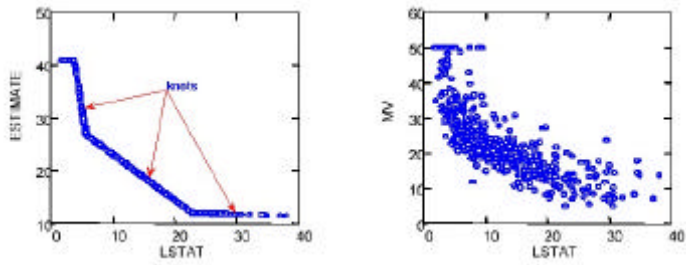

Fig. 2 – MARS data estimation using splines and knots.

Table 2. Test Performance of Different Neural Network Training Functions

| Function | No of Epochs | Accuracy (%) |
|---|---|---|
| Gradient descent | 3500 | 61.70 |
| Gradient descent with momentum | 3500 | 51.60 |
| Adaptive learning rate | 3500 | 95.38 |
| Resilient back propagation | 67 | 97.04 |
| Fletcher-Reeves conjugate gradient | 891 | 82.18 |
| Polak-Ribiere conjugate gradient | 313 | 80.54 |
| Powell-Beale conjugate gradient | 298 | 91.57 |
| Scaled conjugate gradient | 351 | 80.87 |
| BFGS quasi-Newton method | 359 | 75.67 |
| One step secant method | 638 | 93.60 |
| Levenberg-Marquardt | 17 | 76.23 |
| Bayesian regularization | 533 | 64.15 |

Table 3. Performance of the Best Neural Network Training Function (Resilient Back Propagation)

| Class of Attack | Normal | Probe | DoS | U2Su | R2L | % |
|---|---|---|---|---|---|---|
| Normal | 1394 | 5 | 1 | 0 | 0 | 99.6 |
| Probe | 49 | 649 | 2 | 0 | 0 | 92.7 |
| DoS | 3 | 101 | 4096 | 2 | 0 | 97.5 |
| U2Su | 0 | 1 | 8 | 12 | 4 | 48.0 |
| R2L | 0 | 1 | 6 | 21 | 535 | 95.0 |
| % | 96.4 | 85.7 | 99.6 | 34.3 | 99.3 | |

Table 4. Test Performance of SVMs

| Class of Attack | Training Time (sec) | Testing Time (sec) | Accuracy (%) |
|---|---|---|---|
| Normal | 7.66 | 1.26 | 99.55 |
| Probe | 49.13 | 2.10 | 99.70 |
| DoS | 22.87 | 1.92 | 99.25 |
| U2Su | 3.38 | 1.05 | 99.87 |
| R2L | 11.54 | 1.02 | 99.78 |

Table 5. Performance Comparison of Testing for 5 class Classifications

| Class of Attack | Accuracy (%) | | | | |
|---|---|---|---|---|---|
| | SVM | RP | SCG | OSS | MARS |
| Normal | 98.42 | 99.57 | 99.57 | 99.64 | 96.08 |
| Probe | 98.57 | 92.71 | 85.57 | 92.71 | 92.32 |
| DoS | 99.11 | 97.47 | 72.01 | 91.76 | 94.73 |
| U2Su | 64 | 48 | 0 | 16 | 99.71 |
| R2L | 97.33 | 95.02 | 98.22 | 96.80 | 99.48 |

We used the same testing data (6890), same network architecture and same activations functions to identify the best training function that plays a vital role for in classifying intrusions. Table 2 summarizes the performance of the different learning algorithms. The top-left entry of Table 3 shows that 1394 of the actual "normal" test set were detected to be normal; the last column indicates that 99.6 % of the actual "normal" data points were detected correctly. In the same way, for "Probe" 649 of the actual "attack" test set were correctly detected; the last column indicates that 92.7% of the actual "Probe" data points were detected correctly. The bottom row shows that 96.4% of the test set said to be "normal" indeed were "normal" and 85.7% of the test set classified, as "probe" indeed belongs to Probe. The overall accuracy of the classification is 97.04 with a false positive rate of 2.76% and false negative rate of 0.20.

## 5.3 SVM Experiments

The data set described in section 5 is being used to test the performance of support vector machines. Because SVMs are only capable of binary classifications, we will need to employ five SVMs, for the 5-clas classification problem in intrusion detection, respectively. We partition the data into the two classes of "Normal" and "Rest" (Probe, DoS, U2Su, R2L) patterns, where the Rest is the collection of four classes of attack instances in the data set. The objective is to separate normal and attack patterns. We repeat this process for all classes. Training is done using the RBF (radial bias function) kernel option; an important point of the kernel function is that it defines the feature space in which the training set examples will be classified. Table 4 summarizes the results of the experiments.

## 6 CONCLUSIONS

A number of observations and conclusions are drawn from the results reported: MARS is superior to SVMs in respect to classifying the most important classes (U2Su and R2L) in terms of the attack severity. SVMs outperform ANNs in the important respects of scalability (the former can train with a larger number of patterns, while would ANNs take a long time to train or fail to converge at all when the number of patterns gets large); training time and running time (SVMs run an order of magnitude faster); and prediction accuracy. SVMs easily achieve high detection accuracy (higher than 99%) for each of the 5 classes of data, regardless of whether all 41 features are used, only the important features for each class are used, or the union of all important features for all classes are used. Resilient back propagation achieved the best performance among the neural network learning algorithms in terms of accuracy (97.04 %) and faster convergence (67 epochs). We note, however, that the difference in accuracy figures tend to be very small and may not be statistically significant, especially in view of the fact that the 5 classes of patterns differ in their sizes tremendously. More definitive conclusions can only be made after analysing more comprehensive sets of network traffic data.

Finally, another gifted research line includes the potential use of MARS hybridized with self-organized ant-like evolutionary models as proposed in past works (Ramos, 2003; Abraham, 2003). The implementation of this swarm intelligence along with *Stigmergy* (Ramos, 2002) and the study of ant colonies behaviour and their self-organizing capabilities are decisively of direct interest to knowledge retrieval/management and decision support systems sciences. In fact they can provide new models of distributed, adaptive and collective organization, enhancing MARS data estimation on ever changing environments (e.g. dynamic data on real-time), as those we now increasingly tend to face over new disseminated Information Systems paradigms and challenges.

## ACKNOWLEDGEMENTS

Support for this research received from *ICASA* (Institute for Complex Additive Systems Analysis, a division of *New Mexico Tech*), *U.S. Department of Defense IASP* and *NSF* capacity building grant is gratefully acknowledged, as well as for *FCT PRAXIS XXI* research fellowship, *Science & Technology Foundation* - Portugal. Finally, we would also like to acknowledge many insightful conversations with Dr. *Jean-Louis Lassez* and *David Duggan* that helped clarify some of our ideas.